\documentclass[letterpaper, 10 pt, conference]{ieeeconf}  

\IEEEoverridecommandlockouts                              

\usepackage[utf8]{inputenc}
\usepackage{graphicx}

\title{\LARGE \bf
FINDING AND FOLLOWING OF HONEYCOMBING REGIONS IN COMPUTED TOMOGRAPHY LUNG IMAGES BY DEEP LEARNING}

\author{Emre EĞRİBOZ$^{1}$, Furkan KAYNAR$^{1}$, Songül VARLI$^{1}$, Benan MÜSELLİM$^{2}$, Tuba SELÇUK$^{3}$
\thanks{$^{1}$ Yıldız Technical University,	Computer Engineering Department
        {\tt\small \{emre.egriboz, furkankaynary\}@gmail.com, songul@ce.yildiz.edu.tr}}%
\thanks{$^{2}$İstanbul University, Cerrahpaşa Faculty of Medicine, Department of Chest Diseases 
        {\tt\small benanmusellim@gmail.com}}%
\thanks{$^{3}$Haseki Education Research Hospital, Department of Radiology
        {\tt\small drtubas@gmail.com}}%
}

\begin{document}

\maketitle
\thispagestyle{empty}
\pagestyle{empty}

\begin{abstract}
In recent years, besides the medical treatment methods in medical field, Computer Aided Diagnosis (CAD) systems which can facilitate the decision making phase of the physician and can detect the disease at an early stage have started to be used frequently. The diagnosis of Idiopathic Pulmonary Fibrosis (IPF) disease by using CAD systems is very important in that it can be followed by doctors and radiologists. It has become possible to diagnose and follow up the disease with the help of CAD systems by the development of high resolution computed imaging scanners and increasing size of computation power.
The purpose of this project is to design a tool that will help specialists diagnose and follow up the IPF disease by identifying areas of honeycombing and ground glass patterns in High Resolution Computed Tomography (HRCT) lung images. Creating a program module that segments the lung pair and creating a deep learning model from given Computed Tomography (CT) images for the specific diseased regions thanks to doctors are the main purposes of this work. Through the created model, program module will be able to find special regions in given new CT images. 
In this study, the performance of lung segmentation was tested by the Sørensen-Dice coefficient method and the mean performance was measured as 90.7\%, testing of the created model was performed with data not used in the training stage of the Convolutional Neural Network (CNN), and the average performance was measured as 87.8\% for healthy regions, 73.3\% for ground-glass areas and 69.1\% for honeycombing zones.
\end{abstract}

\section{INTRODUCTION}
\label{sec:intro}
\begin{figure}
\centering
\includegraphics[width=0.25\textwidth]{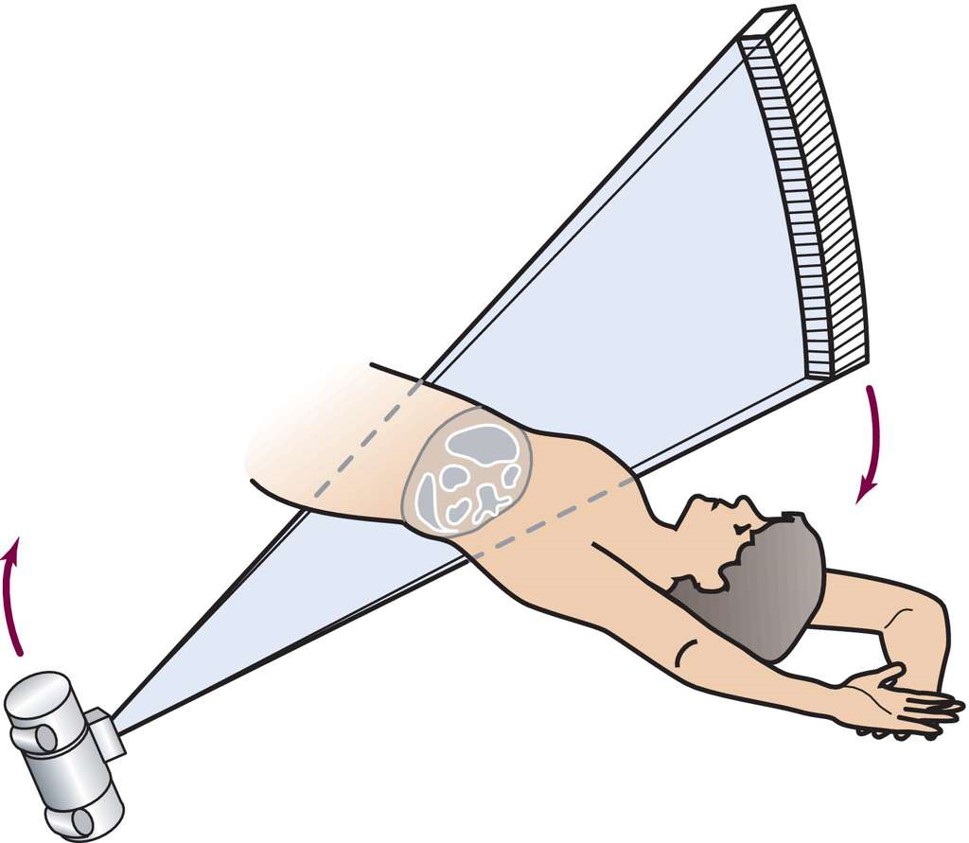}
\caption{Tomography Capturing Step}
\label{fig:tomography_device}
\end{figure}

CT is a three-dimensional cross-sectional imaging method based on an X-ray machine. It tries to display the internal structure of an object with the rays it sends from different angles. An image from the capture moment of the tomography device is presented in Figure \ref{fig:tomography_device}.
In years of development, the CT scans have become faster and have started to generate higher resolution images. In this way, the diagnosis and treatment of the doctors started to produce more efficient results.
The need for accurate and consistent analysis of high resolution and multiple sections of tomography images, which is the output of modern tomography devices, increases.
Computer Aided Diagnostic (CAD) systems are produced to meet this analysis need.

Deep Learning is an artificial neural network algorithm that simulates the functioning of the human brain in the processing of data and creates models for decision making. In this method, the decision model is created by training and verification phases. The amount of data used in these phases, cleaning and distribution according to a statistical model are important parameters for the accuracy of the model to be formed. CNNs are artificial networks that are often formed from image patterns and are a special study area of deep learning. 

\begin{figure}[htb]
\begin{minipage}[b]{.54\linewidth}
  \centering
  \centerline{\includegraphics[width=4.5cm]{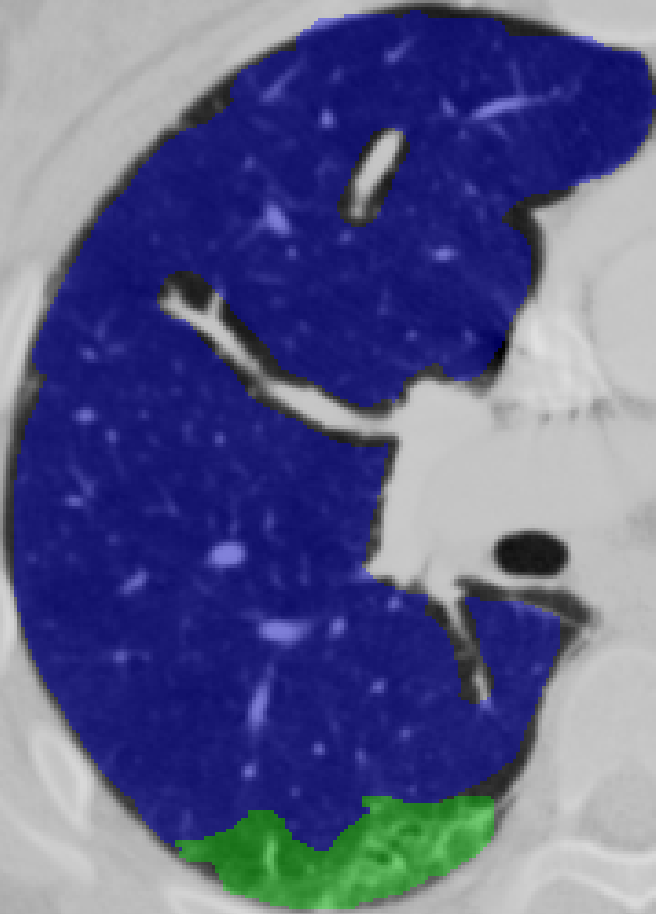}}
  \centerline{(a) Ground-glass pattern}\medskip
\end{minipage}
\hfill
\begin{minipage}[b]{0.44\linewidth}
  \centering
  \centerline{\includegraphics[width=3.82cm]{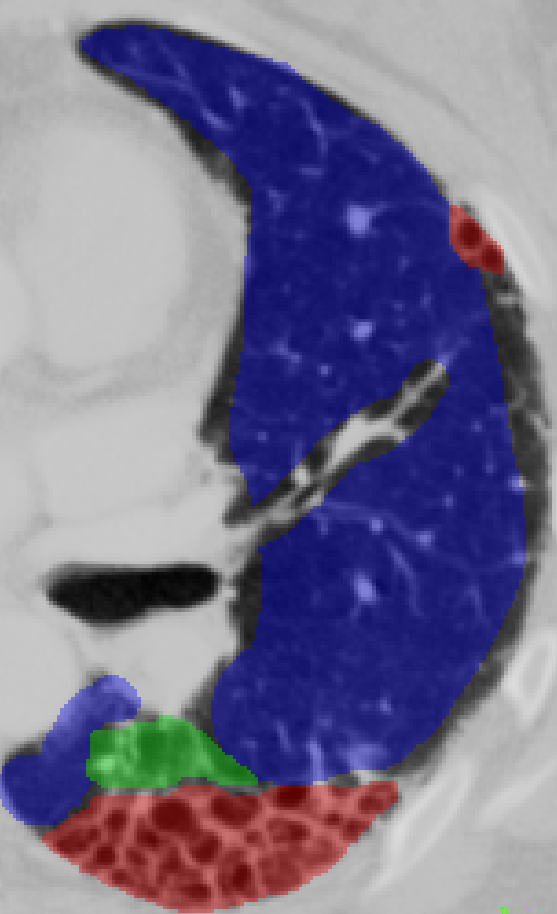}}
  \centerline{(b) Honeycombing pattern}\medskip
\end{minipage}
\caption{(a) Ground-glass pattern which is colored green is seen in the posterior region of the right lob of lung. (b) Honeycombing pattern which is colored red is seen in the entire posterior region of left lob of lung.}
\label{fig:honeycombing_groundglass}
\end{figure}

IPF which is a type of Interstitial Lung Disease (ILD), is a progressive lung disease that hardens the lungs and makes the patient increasingly breathing difficult. Fibrosis is seen as scar tissue or a hard collagen surrounding the alveoli. The diagnosis is made based on the typical pattern of lung tomography, which is called as honeycombing region. Also, the typical pattern of the early period of lung fibrosis is called as ground-glass region. Tomography images of the honeycombing and ground-glass regions are presented in Figure \ref{fig:honeycombing_groundglass}.

Due to the high number of sections, small areas or difficult to diagnose, the detection of specific areas of IPF disease on tomography images and the process-related follow-up of disease progression have become a difficult task for field experts. CT images were processed by image processing methods. Mean Hounsfield Unit (HU) value of all lungs \cite{bib_mean_density_histogram}, density histogram analysis \cite{bib_mean_density_histogram}, density mask technique \cite{bib_density_mask},\cite{bib_threshold}  and tissue classification methods \cite{bib_tissue_classification} were used for the characterization of IPF disease. These studies show the tight correlation between HU values and IPF and the patterns of tissue obtained from CT images are can be successfully used in prediction by its differentiation ability. This project provides a digital system for the detection of the honeycombing and ground-glass pattern by computer, which play a decisive role in the diagnosis of IPF disease on CT images. It is also a CAD project that helps the doctors and radiologists in the diagnosis and follow-up of IPF disease. Unfortunately there are no previous studies to benchmark the algorithms used in the prediction of the regions of disease.


The next chapters of this article are organized as follows: In the second chapter, the characteristics of the dataset,  lung segmentation, the cutting of the lung images and the deep learning model which are the main topics of the system design are mentioned. In the third chapter, the details of the experimental results are explained and in the last section, the results obtained from this study are summarized and future studies are mentioned.

\section{System Design}

\label{sec:systemdesign}

\subsection{Dataset}

CT image data is saved in DICOM (Digital Imaging and Communications in Medicine) file format, which is a international standard for medical imaging, by tomography machines. During all phases, the HU scale which was used to digitize DICOM images is used. 


Special areas on the lungs are divided into 3 groups: ground-glass, honeycombing and healthy. The marking process is performed by any scoring tool by field experts who are generally doctors and radiologists. The dataset has data from 6 patients, per includes from 32 to 170 DICOM slices. The detailed data sizes of the patients are given in Section \ref{subsec:blocking}.
Scoring data is saved in NIFTI (The Neuroimaging Informatics Technology Initiative) file format separately from tomography images, which is a global standard for neuroimaging marking.

\subsection{Lung Segmentation}

Different substances have different HU values. HU values of some items related to our project are provided in Table \ref{table:hounsfield}. Materials containing similar substances have similar HU values. In other words, the hardness of the material could be thought as HU value.

\begin{table}[h!]
\centering
\caption{Hounsfield Unit Scale Values}
    \begin{tabular}{|l|r|}
        \hline
        \textbf{Substance}& \textbf{Value(HU)} \\ \hline
        Air & -1000  \\  \hline
        Lung & -700 to -600  \\  \hline
        Water & 0  \\  \hline
        Blood & 13 to 50  \\  \hline
        Kidney & 20 to 45  \\  \hline
        Bone & 200 to 3000  \\  \hline
        Gold & 30000  \\  \hline
        
    \end{tabular}
    \label{table:hounsfield}
\end{table}

\begin{figure}[htb]
\begin{minipage}[b]{.48\linewidth}
  \centering
  \centerline{\includegraphics[width=4.0cm]{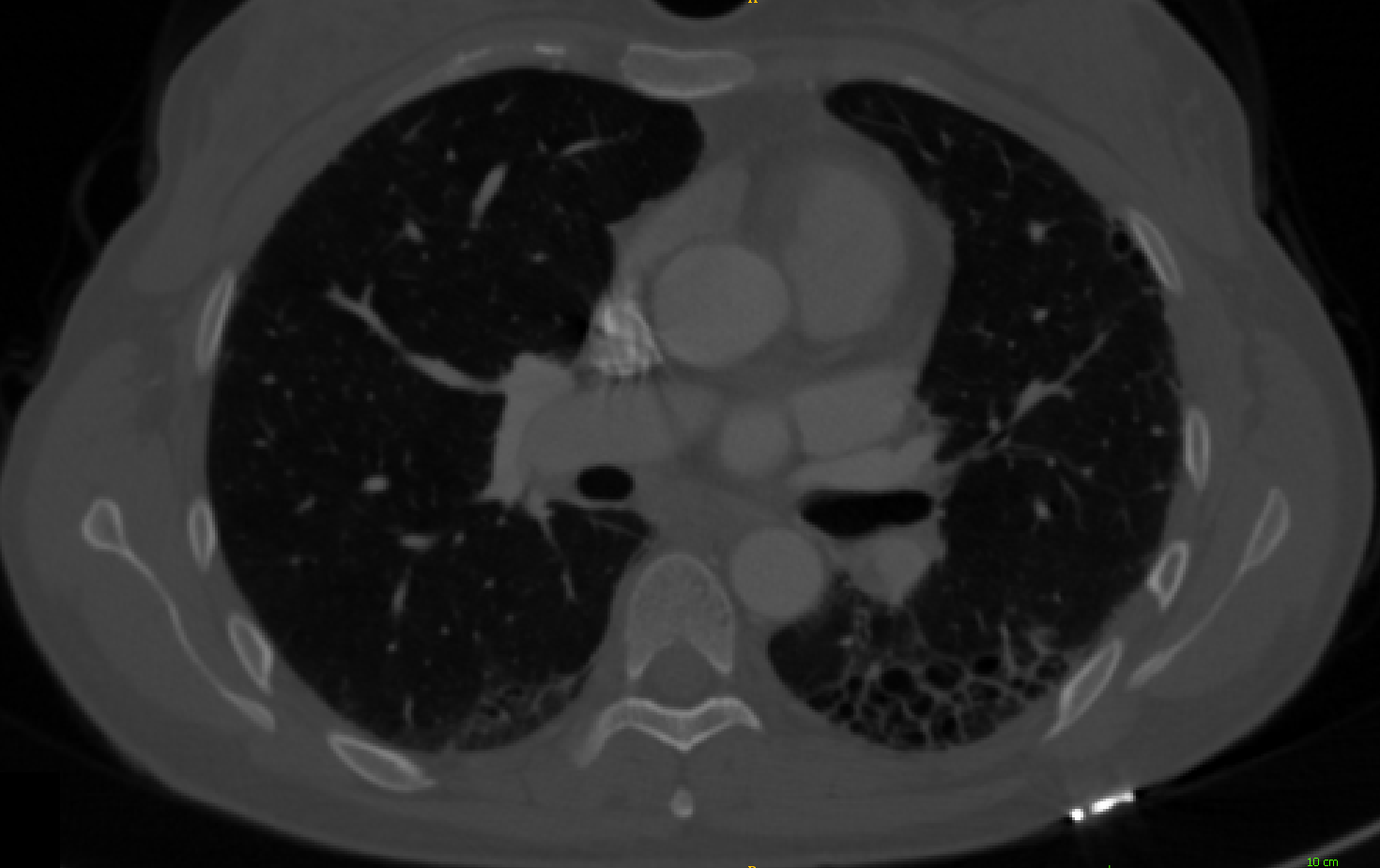}}
  \centerline{(a) Original CT image}\medskip
\end{minipage}
\hfill
\begin{minipage}[b]{0.48\linewidth}
  \centering
  \centerline{\includegraphics[width=4.0cm]{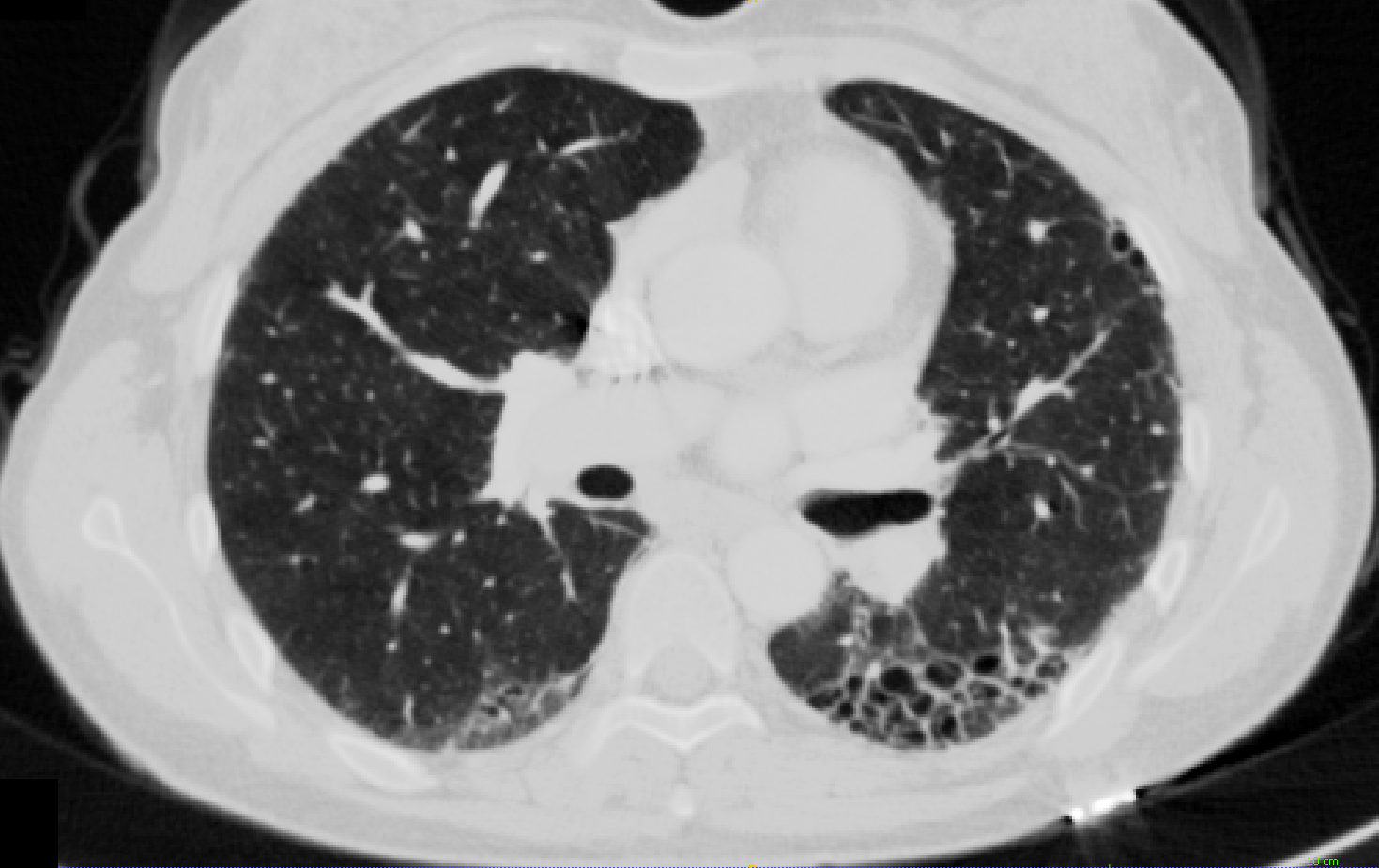}}
  \centerline{(b) Enriched CT image}\medskip
\end{minipage}
\caption{Enrichment stage sample.}
\label{fig:enrichment_stage}
\end{figure}

Before the lung segmentation phase begins the CT images are enriched to work more clearly and human-viewable. The benefit of the enrichment stage can be seen in Figure \ref{fig:enrichment_stage}.

\begin{figure}[htb]
\begin{minipage}[b]{.48\linewidth}
  \centering
  \centerline{\includegraphics[width=4.0cm]{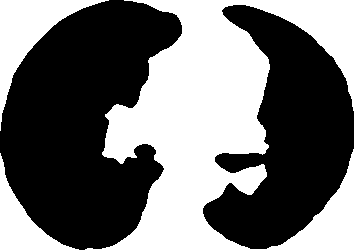}}
  \centerline{(a) Obtained mask}\medskip
\end{minipage}
\hfill
\begin{minipage}[b]{0.48\linewidth}
  \centering
  \centerline{\includegraphics[width=4.0cm]{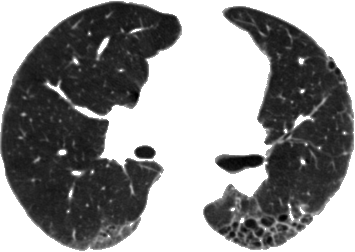}}
  \centerline{(b) Masked lung pair}\medskip
\end{minipage}
\caption{Lung segmentation result sample.}
\label{fig:mask_stage}
\end{figure}

The lungs are all filled with air. Then the necessary mask can be gotten by thresholding at the lung by HU value(-700 to -600) and the air HU value(-1000) to segment the lung pair. Then, the mask is obtained by removing small objects on lungs. After the obtained mask is applied to the tomography image containing the original HU values, lung segmentation is completed. Figure \ref{fig:mask_stage} shows mask that will be used to segment current section and shows also masked lung pair.

\subsection{Blocking}
\label{subsec:blocking}
\begin{figure}
\centering
\includegraphics[width=0.48\textwidth]{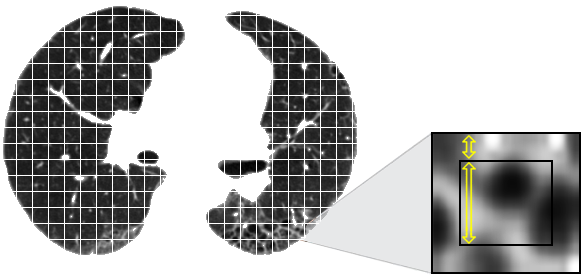}
\caption{Blocking process: Short arrow represents padding length, long arrow represents ROI length.}
\label{fig:pixel_raster}
\end{figure}

Segmented lung images are need to be partitioned into blocks and presented as input to the network. Blocks are cropped by the region of interest (ROI) and padding dimensions. The blocks are shifted to get a new ROI. ROI regions are not overlapped. The dominant marked output of the ROI region for the cropped part is considered to be the output for the whole part. ROI and padding sizes, cropped blocks and grid are presented in Figure \ref{fig:pixel_raster}. 

All experiments in the study used 4 pixels as the ROI and 4 pixels as the padding. According to these ROI and padding length, the number of blocks obtained from the sections are given in Table \ref{table:detaileddatasizes}. The number of blocks of regions is proportioned to the lowest region' s sample count to avoid the over-sampling. For this reason, the number of total blocks of some patients is low compared to the number of sections.

\begin{table}[h!]
\centering
\caption{Number of Blocks by Patient and Region}
    \resizebox{\linewidth}{!}{%
    \begin{tabular}{|l|r|r|r|r|}
        \hline
         & \textbf{\#of Section}& \textbf{Honeycombing}& \textbf{G. Glass}& \textbf{Healthy} \\ \hline
        1$^{st}$ Patient & 36  & 10.204 & 15.304 & 25.507  \\  \hline
        2$^{nd}$ Patient & 32  & 15.174 & 16.705 & 21.317  \\  \hline
        3$^{rd}$ Patient & 41  & 393 & 588 & 980  \\  \hline
        4$^{th}$ Patient & 170  & 26.778 & 40.165 & 66.942  \\  \hline
        5$^{th}$ Patient & 159  & 28.441 & 18.962 & 47.402  \\  \hline
        6$^{th}$ Patient & 39  & 1.366 & 2.047 & 3.412  \\  \hline
        \textbf{Total} & \textbf{447}  & \textbf{82.356} &\textbf{93.741}& \textbf{165.560} \\  \hline
        
    \end{tabular}
    }
    \label{table:detaileddatasizes}
\end{table}

\subsection{Deep Learning Model}
Deep learning methods have been widely used in recent years with its high successes especially in image classification \cite{bib_image_recognization}. CNN is one of the deep learning algorithm used by LeCun for character recognition for the first time\cite{bib_conv_neu_netw}. CNN was applied in small size data at the time when it was first proposed. The development of hardware technologies, it has been tried to be applied in large scale data.

CNN models uses different kinds of layers. The layers used in this study are dense layer which consists regular neurons with their weights, 2 dimensional convolutional layer which extracts features by applying spatial convolution over images, 2 dimensional max-pooling layer which does down-sampling over images by reducing their dimensionality because of memory limitation, dropout layer which prevents the overfitting by dropping specified percentage of dense layer. Activation functions are used to draw the results from the neurons to a common range.

\begin{figure*}
\centering
\includegraphics[width=\textwidth]{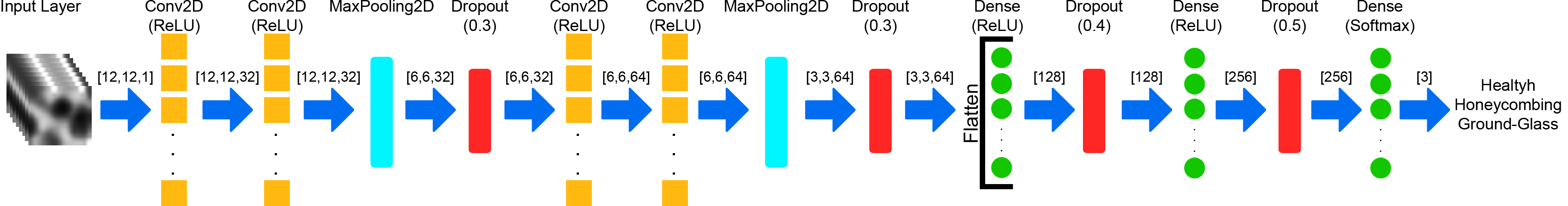}
\caption{Structure of deep learning network model}
\label{fig:network}
\end{figure*}

The success of CNN depends on the number of filters used, the depth of the network and the simplifications performed in the pool layers in the structured architecture. Therefore, there are different CNN architectures according to different problem types. The proposed model is depicted in Figure \ref{fig:network}. 

\section{Experimental Results}
\label{sec:experimentalresults}

Experimental results include the success rate of lung segmentation and the success rate of predicting the diseased regions.

Experiments in this study were applied on Windows-10 operating system and Keras application program interface which uses Google TensorFlow framework in backend by using Python programming language.
The computer runs on that system is powered by a 4-core Intel i7-6700HQ with 2.6 GHz 6MB cache processor, 16GB of memory and a NVIDIA GeForce GTX950M model with 4GB memory graphical processor.
\subsection{Lung Segmentation Accuracy}
The performance in the lung segmentation phase was measured by Sørensen–Dice coefficient. 
The Dice similarity coefficient specifies a similarity ratio for the segmented lung by analytical method segmentation and professional segmentation between 0 and 1. 0 indicates that two images do not intersect at any point, and 1 indicates full intersection for all points.

\begin{equation}
\label{eqn:sorensen}
    Dice(A,B)\ = \frac{2 * | intersection(A,B) |}{ | A | + | B | }  
\end{equation}

The Dice similarity coefficient of two sets A and B is expressed as in (\ref{eqn:sorensen}) where \textbar A\textbar \ represents the cardinal of set A. Lung segmentation accuracy by patient samples are provided in Table \ref{table:lungsegmentationaccurraccy}.

\begin{table}[h!]
\centering
\caption{Lung Segmentation Accuracy by Patient}
    \begin{tabular}{|l|r|r|}
        \hline
         & \textbf{\#of Section}& \textbf{Dicemetric(\%)} \\ \hline
        1$^{st}$ Patient & 36  & 0.920  \\  \hline 
        2$^{nd}$ Patient & 32  & 0.916  \\  \hline 
        3$^{rd}$ Patient & 41  & 0.903  \\  \hline   
        4$^{th}$ Patient & 170  & 0.845  \\  \hline   
        5$^{th}$ Patient & 159  & 0.980  \\  \hline   
        6$^{th}$ Patient & 39  & 0.871  \\  \hline   
        \textbf{Weighted Average} & \textbf{-} & \textbf{0.907} \\  \hline
        
    \end{tabular}
    \label{table:lungsegmentationaccurraccy}
\end{table}

\subsection{Region Prediction Accuracy}
Prediction performance is measured by one patient out method. In the one patient out method, each data is used once as training and once as test data. The accuracy of the model is calculated by taking the weighted average of the accuracy values according to the number of sections. Region prediction precision table by patient samples are provided in Table \ref{table:predictionaccurraccy}. 

The data of 3$^{rd}$ patient could not be used because the tomography images were not clear due to the patient's motion inside the tomography device so patterns are broken, up to negatively affect the network' s weights. Higher accuracy was obtained for both ground-glass and honeycombing regions in the 4$^{th}$ and 5$^{th}$ patients whose CT has more sections than others. The worst result was measured as 40\% in ground-glass accuracy of 6$^{th}$ patient. 

In general, the system have some trouble distinguishing between ground-glass and honeycombing fields and confuses them. It has been difficult to differentiate them for even doctors and radiologists. The system has obtained 87\% accuracy in the detection of healthy areas.


\begin{table}[h!]
\centering
\caption{Region Prediction Accuracy by Patient}
    \resizebox{\linewidth}{!}{%
    \begin{tabular}{|l|r|r|r|r|}
        \hline
         & \textbf{\#of Section}& \textbf{Honeycombing(\%)}& \textbf{G. Glass(\%)}& \textbf{Healthy(\%)} \\ \hline
        1$^{st}$ Patient & 36  & 0.601 & 0.606 & 0.640  \\  \hline
        2$^{nd}$ Patient & 32  & 0.601 & 0.906 & 0.508  \\  \hline
        3$^{rd}$ Patient & 41  & - & - & -  \\  \hline
        4$^{th}$ Patient & 170  & 0.631 & 0.801 & 0.952  \\  \hline
        5$^{th}$ Patient & 159  & 0.756 & 0.736 & 0.906  \\  \hline
        6$^{th}$ Patient & 39  & 0.846 & 0.401 & 0.970  \\  \hline
        \textbf{W. Average} & \textbf{-}  & \textbf{0.691} &\textbf{0.733}& \textbf{0.878} \\  \hline
        
    \end{tabular}
    }
    \label{table:predictionaccurraccy}
\end{table}

\section{Results And Discussion}
\label{sec:results}
The purpose of this project is to design a tool that will help specialists diagnose and follow up the disease by identifying areas of honeycombing and ground glass areas in high-resolution lung images. Creating a program module that segments the lung pair and creating a deep learning model from given CT files for the special diseased regions thanks to doctors are the main purposes of this work. Through the created model, program module will be able to find special regions in given new CT files. 

In future studies, different additional image processing methods such as morphological operations can be applied to lung segmentation phase. By adding more patients to the system, it is hoped that success in the deep learning model will increase further. Also, additional features can be manually extracted and the channels in the input can be increased to improve the weights in the network. With such improvements, the performance of both segmentation and prediction can be increased.


\begin{thebibliography}{00}

\bibitem{bib_mean_density_histogram} Murray J. Gilman, Richard G. Laurens, James W. Somogyi and Eric G. Honig, "CT Attenuation Values of Lung Density in Sarcoidosis" Journal of Computer Assisted Tomography, 7(3):407–410, JUN 1983

\bibitem{bib_density_mask}Nestor L.Müller, Roberta R.Miller and Raja T.Abboud, "An Objective Method to Quantitate Emphysema Using Computed Tomography", Chest, Volume 94, Issue 4, October 1988, Pages 782-787

\bibitem{bib_threshold}Wang, Z., Gu, S., Leader, J.K. et al., "Optimal threshold in CT quantification of emphysema", Eur Radiol (2013) 23: 975

\bibitem{bib_tissue_classification}Ye Xu, Edwin J.R. van Beek, Yu Hwanjo, Junfeng Guo, Geoffrey McLennan and Eric A. Hoffma, "Computer-aided Classification of Interstitial Lung Diseases Via MDCT: 3D Adaptive Multiple Feature Method (3D AMFM)", Academic Radiology, Volume 13, Issue 8, August 2006, Pages 969-978

\bibitem{bib_conv_neu_netw}LeCun,  Yann,  et  al.  "Gradient-based  learning  applied  to  document recognition.", Proceedings of the IEEE 86.11 (1998): 2278-2324


\bibitem{bib_image_recognization} K. He, X. Zhang, S. Ren, and J. Sun, “Deep residual learning for image recognition,” in Proceedings of the IEEE Conference on Computer Vision and Pattern Recognition, 2016, pp. 770–778.


\end{thebibliography}
\end{document}